\documentclass[11pt]{article}

\usepackage[utf8]{inputenc}
\usepackage{amsmath,amsfonts,amssymb,amsthm}
\usepackage{graphicx,psfrag}
\usepackage{color}
\usepackage[ruled]{algorithm2e}
\usepackage{url}
\usepackage[normalem]{ulem}
\usepackage{paralist}
\usepackage{enumitem}

\newcommand{\N}{\mathbb{N}}

\newcommand{\bmat}{\begin{pmatrix}}
\newcommand{\emat}{\end{pmatrix}}

\title{Limits of End-to-End Learning}
\author{Tobias Glasmachers\\
		Institute for Neural Computation\\
		Ruhr-University Bochum, Germany\\
		\texttt{tobias.glasmachers@ini.rub.de}}
\date{}

\begin{document}

\maketitle

\begin{abstract}
End-to-end learning refers to training a possibly complex learning
system by applying gradient-based learning to the system as a whole.
End-to-end learning system is specifically designed so that all modules
are differentiable. In effect, not only a central learning machine, but
also all ``peripheral'' modules like representation learning and memory
formation are covered by a holistic learning process. The power of
end-to-end learning has been demonstrated on many tasks, like playing a
whole array of Atari video games with a single architecture. While
pushing for solutions to more challenging tasks, network architectures
keep growing more and more complex.

In this paper we ask the question whether and to what extent end-to-end
learning is a future-proof technique in the sense of \emph{scaling} to
complex and diverse data processing architectures. We point out
potential inefficiencies, and we argue in particular that end-to-end
learning does not make optimal use of the modular design of present
neural networks. Our surprisingly simple experiments demonstrate these
inefficiencies, up to the complete breakdown of learning.
\end{abstract}

\section{Introduction}

We are today in the position to train rather deep and complex neural
networks in an \emph{end-to-end} (e2e) fashion, by gradient descent. In
a nutshell, this amounts to scaling up the good old backpropagation
algorithm (see \cite{schmidhuber2015deep} and references therein) to
immensely rich and complex models. However, the end-to-end learning
philosophy goes one step further: carefully ensuring that all modules of
a learning systems are differentiable with respect to all adjustable
parameters (weights) and training this system as a whole are lifted to
the status of principles.

This elegant although straightforward and somewhat brute-force technique
has been popularized in the context of deep learning. It is a seemingly
natural consequence of deep neural architectures blurring the classic
boundaries between learning machine and other processing components by
casting a possibly complex processing pipeline into the coherent and
flexible modeling language of neural networks. The approach yields
state-of-the-art results \cite{collobert2011natural,alexnet,mnih2015human}.
Its appeal is a unified training scheme that makes most of the available
information by taking labels (supervised learning) and rewards
(reinforcement learning) into account, instead of relying only on the
input distribution (unsupervised pre-training).
Excellent recent examples of studies deeply rooted in the e2e learning
philosophy---naming only a few well known ones---are the neural Turing
machine \cite{graves2014neural} and the differentiable neural computer
\cite{graves2016hybrid}, value iteration networks~\cite{tamar2016value},
and vision-based navigation in 3D
environments~\cite{mirowski2016learning}.
These achievements defy the well-known principal limitations of gradient
descent, namely local optima and slow convergence in various
circumstances depending on the exact algorithm in use, typically on
badly conditioned problems. Intuitively, both of these problems may
become more severe when network architectures grow in complexity.

\noindent
The current state-of-the-art is based on a long sequence of
technological advancements, which took only about one decade to evolve.
Major steps and factors are listed in the sequel.
\begin{compactitem}
\item
	Interestingly, the success story of deep learning as we know it
	today (notwithstanding many earlier findings which, however, did not
	make a comparable impact at the time \cite{lecun2005offroad,schmidhuber2015deep})
	started with structured, layer-wise training of deep belief
	networks~\cite{hinton2006fast} and stacked
	auto-encoders~\cite{vincent2008extracting}. These techniques can be
	understood as the exact opposite of e2e learning. Although they mark
	breakthroughs of representation learning, they are widely considered
	to have become obsolete.
\item
	More ``classic'' networks moved back into the focus due to progress
	on efficient GPU implementations of backpropagation for deep
	convolutional neural networks (CNNs) and, as a consequence,
	significant progress on computer vision
	tasks~\cite{cirecsan2011committee}.
\item
	GPU-based processing allowed to scale to extremely large networks
	and learning problems. The availability of huge training sets and
	the computational power to process them were key prerequisites for
	solving previously intractable tasks with deep learning
	techniques~\cite{alexnet}.
\item
	The development went hand-in-hand with tremendous progress in the
	area of stochastic gradient descent (SGD). On the one hand side,
	linearly convergent methods for finite sums were
	developed~\cite{schmid2013minimizing,johnson2013accelerating}, on
	the other hand, effective component-wise online adaptation
	techniques for learning rate parameters resulted in significant
	speed-ups over plain SGD. The new techniques made deep learning far
	more tractable and resulted in even better
	performance~\cite{zeiler2012adadelta,kingma2014adam}.
\item
	Various easy-to-use deep learning software libraries were developed
	\cite{caffe,keras,matconvnet}, many of which are based on powerful
	symbolic computation software packages like theano~\cite{theano} and
	tensorflow~\cite{tensorflow}.
\item
	The race for solving Imagenet classification \cite{deng2009imagenet}
	resulted in deeper and deeper architectures. It lead to
	creative designs~\cite{szegedy2015going}, and also shed new light on
	the utility of good shortcut connections~\cite{he2016deep}.
\item
	The immense progress even allowed to train highly flexible networks
	in a (vision-based) reinforcement learning fashion
	\cite{mnih2015human,silver2016mastering,mirowski2016learning},
	despite the limited and sometimes delayed information contained in
	the reward signal.
\end{compactitem}
In this paper, we take the freedom to extrapolate this development into
the (foreseeable) future. We discuss the implications of applying e2e
learning methods to considerably more complex systems, which we today
envision to be required for solving the ``holy Grale'' of tasks, namely
operating autonomous agents in a human environment.

As such, and although we present experimental results later on, we
consider this work primarily as a position paper. Our central claim is
that e2e learning has limitations that will keep us from using it in its
current form as the sole way of training networks in the future. This
claim necessarily remains unproven. Also, although we point into a
direction of a possible solution, we do by no means claim to have an
alternative to e2e learning ready for use.

The remainder of this paper is organized as follows. In the next section
we present and discuss the problem in an informal way and point towards
a possible route to solution. We summarize this discussion by collecting
merits and limitations of e2e learning for better overview. Then we
underpin our arguments experimentally: we show how e2e learning can fail
for the training even of rather small systems due to non-trivial
couplings that originate either from the network structure itself or
from the task. We close with a brief conclusion.

\section{Model Engineering vs.\ End-to-end Learning}

State-of-the-art e2e learning enjoys a high degree of automation,
however, it does not work fully autonomously: the design and
configuration of a learning system suited for a given task requires a
certain level of experience and machine learning knowledge. Today's deep
learning systems are engineered from modules, often organized into
layers and groups of layers with specific roles. Prominent examples are
alternating sequences of convolution and pooling layers, fully connected
layers, auto-encoders (bottleneck layers), and LSTM layers. Even
dropout, which is a regularization technique and not a model component,
it often thought of as a layer, which (at least in software) can be
added on top of any other layer. The general proceeding is to compose a
system with the capacity---in principle and by design---to exhibit a
desired behavior.
After the design phase, full automation takes over: weights are
initialized and random, and they are subsequently trained, e2e, with
variants of stochastic gradient descent.

\begin{figure}
\begin{center}
{
	\includegraphics{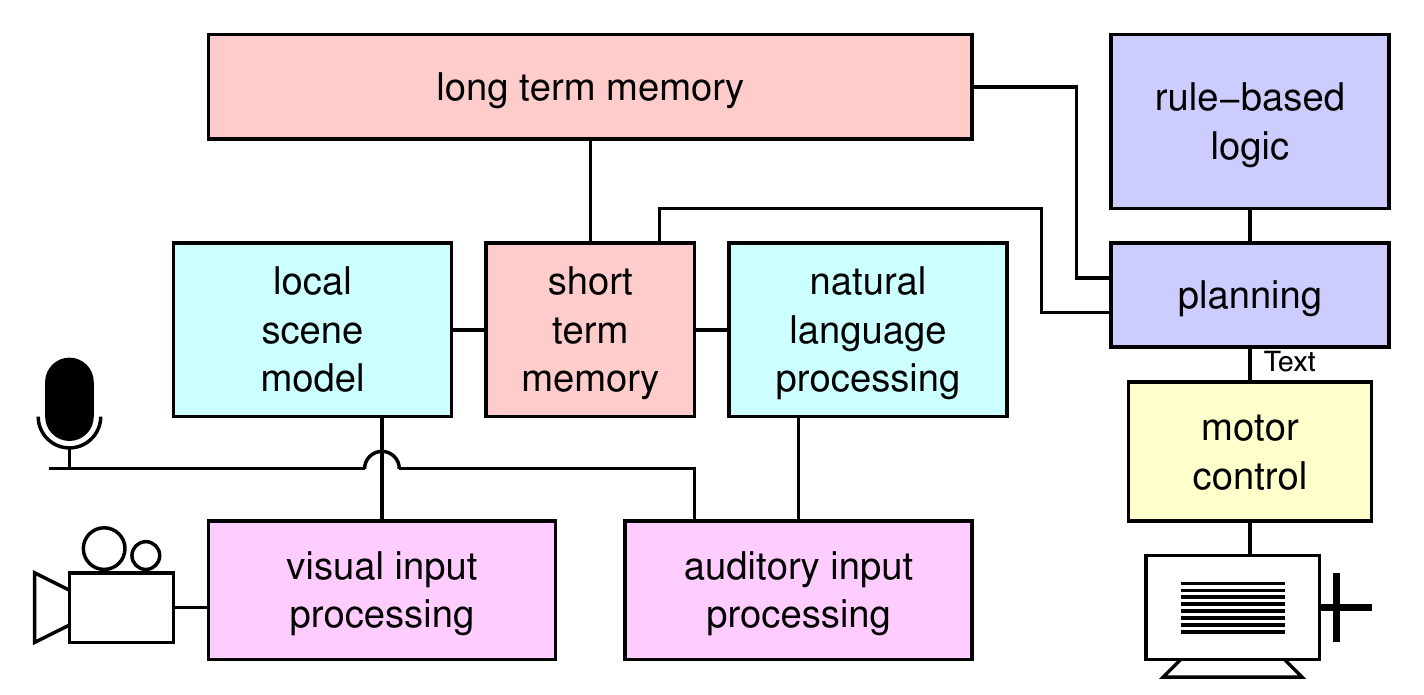}
}
\caption{\label{figure:complexity}
	Fictitious complex learning system composed of a large number of
	modules with diverse roles, and their connectivity. In e2e learning,
	each module is fully differentiable with respect to its parameters.
}
\end{center}
\end{figure}

Driving this to the extreme, one could try to design a full ``brain''
this way, consisting of cortical areas like visual cortex, auditory
cortex, motor cortex, a hippocampus, and so forth. Of course, there is
no need for the design to resemble biological brains in structure and
connectivity, however, for any non-trivial agent we will surely end up
with dedicated modules for sensory data processing, natural language
processing, memory, decision making, planning, motor control, and more
(see figure~\ref{figure:complexity} for an illustration).

Problem decomposition is central to solving a complex problem. This
divide and conquer approach is the core principle of engineering. It is
underlying our network design decisions, and we are well aware that
learning problems become much harder if we don't know how to decompose
them, or if we simply get it wrong.

Hence, divide-and-conquer is heavily used for the \emph{design} of
complex learning machines. However, \emph{training} such a system in an
e2e fashion means \emph{to explicitly ignore the problem decomposition
during learning}.
Instead, training is done in good hope that the structural
preconditioning is sufficiently strong to direct a method as simple as
gradient descent from a random initial state to a highly non-trivial
solution. Intuitively, this is a dangerous assumption: since the
resulting optimization problem is by no means convex, learning can only
work if the task is ``nearly trivial'' given the structure on which it
operates. Data efficiency is a second concern, since unmodeled
interactions between modules may require an amount of training data that
grows possibly exponentially with the number of modules for the problem
to be well-sampled.

Let us return to the above example of the brain-like model. Assume that
each module is realized as a neural network. The roles of these modules
reflect the network designer's intention and his ideas of how to
structure a possible solution to an overwhelmingly complex problem.
However, these roles are not assigned formally to the modules. Nothing
enforces that a specific module actually learns to fulfills its role and
only that role. The role is only suggested by the pre-defined network
structure, and the near-black-box character of complex neural networks
makes even checking whether the assumptions are fulfilled or not a
non-trivial task. Hence, the learning rule may come up with an
alternative solution, or something that looks like a potential solution
during an early training phase. More often than not, this route may lead
to convergence into a poor local optimum.

Against this background it seems obvious that we should respect the
carefully chosen structure during training. There are many possible ways
how to ensure that the training procedure respects our design plan. It
seems that the simplest strategy is to structure the training process
directly in terms of network modules and their connectivity. A simple
procedure with this property is greedy, layer-wise training, however,
its performance is usually inferior to e2e learning due to its inability
to use labels and rewards. We consider finding a training procedure that
makes the powerful of e2e learning approach aware of the network
structure a grand challenge of neural network research.

The insight that existing learning methodologies may not scale to
arbitrarily complex tasks and to learning systems matching this
complexity is by no means new. It motivated various proposals that were
popularized in the domains of control, reinforcement learning, and
artificial intelligence
\cite{ring1994continual,abbeel2004apprenticeship,schmidhuber2004optimal,schmidhuber2010formal}.
A prominent idea is to bootstrap the learning process by solving simple
tasks first, in the hope that solution components of generic value
evolve, which turn out to be helpful building blocks for more complex
tasks later on. All of this can, but does not have to happen in an e2e
fashion. The approach is very different and actually orthogonal to our
vision of organizing the training process of neural networks along their
structure.

\section{Merits and Limitations of End-to-end Learning}

Let us have a more systematic look at the strong and weak spots of the
e2e learning principle. The following lists are intended to provide a
hopefully useful overview of pros and cons of the method, as we see
them, without claiming completeness. The purpose is to offer a basis for
supporting the decision whether e2e learning may be suitable for a
problem at hand or not. It is not our intent to judge the method
according to the number or weight of items on either side.
\paragraph{Merits}
\begin{itemize}[leftmargin=1em]
\setlength\itemsep{0em}
\item
	Conceptual and mathematical beauty: the system is trained in a
	holistic manner based on a single principle.
\item
	Every learning step is directed at the final goal, encoded by
	the overall objective function. There is no need to train
	modules on an ``auxiliary'' objective, unrelated to the task,
	like contrastive divergence or reconstruction error. However, it
	should be noted that e2e learning can in principle incorporate
	auxiliary objectives, possibly affecting only parts of the
	network, usually with the goal of avoiding slow learning and bad
	local optima~\cite{mirowski2016learning}.
\item
	The power of e2e learning for training powerful predictors was
	proven many times in various domains
	\cite{collobert2011natural,alexnet,mnih2015human,silver2016mastering}.
\item
	E2e learning is nicely consistent with the general approach of
	machine learning to take the human expert out of the loop and to
	solve problems in a purely data-driven manner.
\end{itemize}
\paragraph{Limitations}
\begin{itemize}[leftmargin=1em]
\setlength\itemsep{0em}
\item
	The principal limitations of (stochastic) gradient descent
	apply, like slow convergence on ill-conditioned problems and
	convergence into possibly poor local optima.
\item
	In some cases, the learning signals provided by e2e learning can
	be inappropriate. For example, a network module responsible for
	visual representation learning can be trained together with a
	very different module representing a policy \cite{mnih2015human},
	based on possibly sparse and delayed rewards. It seems to be far
	more efficient to train the vision module independently, either
	with unsupervised methods, or to start from one of the many
	available pre-trained networks \cite{alexnet}.
\item
	The valuable information contained problem decomposition that
	resulted in a specific network design is ignored during e2e
	training. Intuitively, this can be dangerous, in particular if
	modules interact in a non-trivial way, since they can hamper
	each other's learning progress. Examples in
	section~\ref{section:experiments} demonstrate that such
	interactions can slow down learning significantly, to the point
	of a complete breakdown.
	It seems that, at least in principle, the decomposition into
	modules could be used to device structured training schemes with
	the potential to overcome these problems.
\end{itemize}
\paragraph{Trade-Offs}
\begin{itemize}[leftmargin=1em]
\setlength\itemsep{0em}
\item
	In a general analysis, the convergence speed of gradient-based
	methods is independent of the number of parameters
	\cite{polyak1963gradient}. Therefore it can be expected that
	training all modules of a network at once takes fewer gradients
	(data samples) than training them independently, e.g., in a
	greedy layer-wise or otherwise structured manner. However, this
	argument can be disputed since the learning speed depends on the
	conditioning of the problem, which can well be significantly
	better for sub-problems involving only few modules.
\end{itemize}
This summary does not yield a general conclusion. We leave it as such
and instead turn to concrete learning systems designed to stress the
limits of e2e learning.

\section{Experiments}
\label{section:experiments}

In previous sections we have made seemingly speculative claims about
possible issues resulting from applications of e2e learning to networks
consisting of interacting modules. In this section we will demonstrate
that such problems indeed exist. We aim to validate the following claims
empirically:
\begin{itemize}
\item
	As the network complexity grows, e2e learning can become
	inefficient.
\item
	For too complex networks, e2e learning can fail completely.
\item
	Training modules one at a time can solve this issue.%
\footnote{Caveat: we do not claim to offer a constructive
  alternative to e2e learning. We do not know a way how to train
  modules in isolation that works generically across a large number
  of tasks and could act as a plug-in replacement for e2e learning.}
\end{itemize}
We demonstrate these claims based on two series of experiments.

\subsection{Scalable Stacking}

This first series of experiments relies on a network structure with
freely scalable complexity, i.e., a class of networks with
parameterizable number of modules. It is no the purpose to design a
learning system that solves a realistic, complex task, but to
demonstrate certain effects in an as clean as possible system. Therefore
we aim to keep things plain and simple.

We start with is a rather trivial supervised classification problem with
$B \in \N$ classes. Inputs are one-hot-encoded numbers in the range
$1, \dots, B$, and so are the target outputs. We apply a standard setup,
which is mini-batch gradient descent training with the ada-delta method
\cite{zeiler2012adadelta} minimizing the cross-entropy loss.

While in a real application network modules may have widely varying
structure and connectivity, we define a single module to be applied in a
sequence of arbitrary length (or depth).

\paragraph{Network Module.}
The module takes one-hot encoded numbers in the range $1, \dots, B$ as
input, sends them through a bottleneck layer of size $b \ll B$ with
sigmoidal (logistic) activation function, and propagates the result to
the output layer of size $B$ with softmax activation. The goal of
learning is to represent the identity mapping restricted to one-hot
encoded vectors; for chained modules, the overall chain shall represent
the identity mapping.

For $B = 2^b$ the problem can be solved by a demultiplexer/multiplexer
logic circuit, which translates one-hot-encoded numbers to a binary
representation and back. We verified empirically that networks can
reliably learn solutions even for values of $B$ significantly larger
than $2^b$, e.g., $B = 64$ and $b = 4$. In principle, even as few as
$b=2$ hidden units suffice, for any number $B$ of inputs/outputs. In the
sequel we resort to the rather simple problem with parameters $B = 10$
and $b = 4$. Including the biases, the total number of weights per
network module is as small as $94$. This module fulfills our
requirements of performing a non-trivial computation, and being
stackable without limit.

\paragraph{Stacked Modules.}
In a first experiment we train $N$ network modules stacked on top of
each other. The data set consists of $10$ points, with input and target
corresponding to the $10$ possible one-hot encodings. We train the
network until it has learned the identity mapping, i.e., the
classification error on the training data drops to zero. Note that we
are not interested in the question whether the network generalizes,
which is trivial in this case since the training data cover all
possible one-hot encodings.

\begin{figure}
\begin{center}
{
	\includegraphics{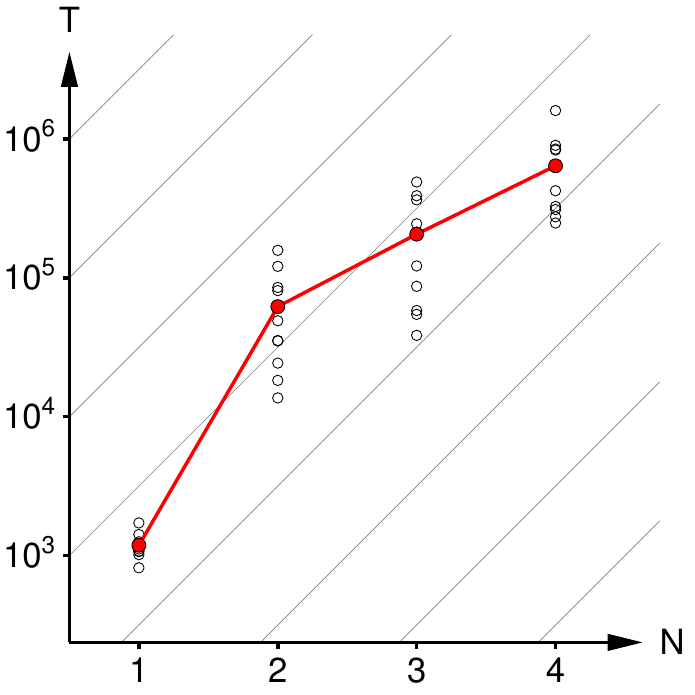}
}
\caption{\label{figure:experiment-1}
	Number of epochs $T$ required for solving the stacked network
	modules task for $N \in \{1, 2, 3, 4\}$ modules. The solid line is
	the averaged over 10 runs, individual runs are represented as dots.
	The gray lines in the background indicate exponential growth of the
	form $T = c \cdot 10^{N}$.
}
\end{center}
\end{figure}

In figure~\ref{figure:experiment-1} we provide the number of epochs
until training is completed. Note that each epoch corresponds to only 10
gradients, therefore the numbers are large. We see a roughly exponential
increase of the training effort. For $N = 5$ modules (not shown), 8 out
of 10 training runs hit the limit of $10^9$ gradients computations
($10^8$ epochs) without solving the task.

In a control experiment, the weights between the different modules were
shared. Hence the module received gradients from all layers at once.
The scaling was similar, but the numbers were even worse.

This is an interesting and insightful result. All networks are rather
small, featuring only a few hundred weights. Training a single module is
easy. However, training multiple modules at once is much harder, with
effort growing exponentially, by roughly one order of magnitude per
module. This implies that training of multiple modules interferes in a
non-trivial and highly disruptive way.

\paragraph{Handwritten Digit Classification.}
We extend the above experiment to test how the difficulty to learn with
multiple modules affects the training of proven components. For this
purpose we created a fairly straightforward network for classifying
handwritten digits from the well-known MNIST data set. The network
consists of the following layers: $5 \times 5$ convolution (relu), $2
\times 2$ max-pooling, $5 \times 5$ convolution (relu), $2 \times 2$
max-pooling, fully connected layer with $200$ nodes (relu), and finally
a fully connected output layer with $10$ nodes (softmax).
We refer to this rather plain architecture as the ``basic MNIST
module''. It achieves non-trivial results ($\approx 99.3\%$ correct
after few epochs), and it could be tweaked further with little effort,
e.g., by adding dropout. Then we stack $N \geq 0$ of the above described
modules on to of this basic MNIST module. The network is trained in an
e2e manner. Training is stopped as soon as the validation error stalls
for at least $20$ epochs in a row.

\begin{figure}
\begin{center}
{
	\includegraphics{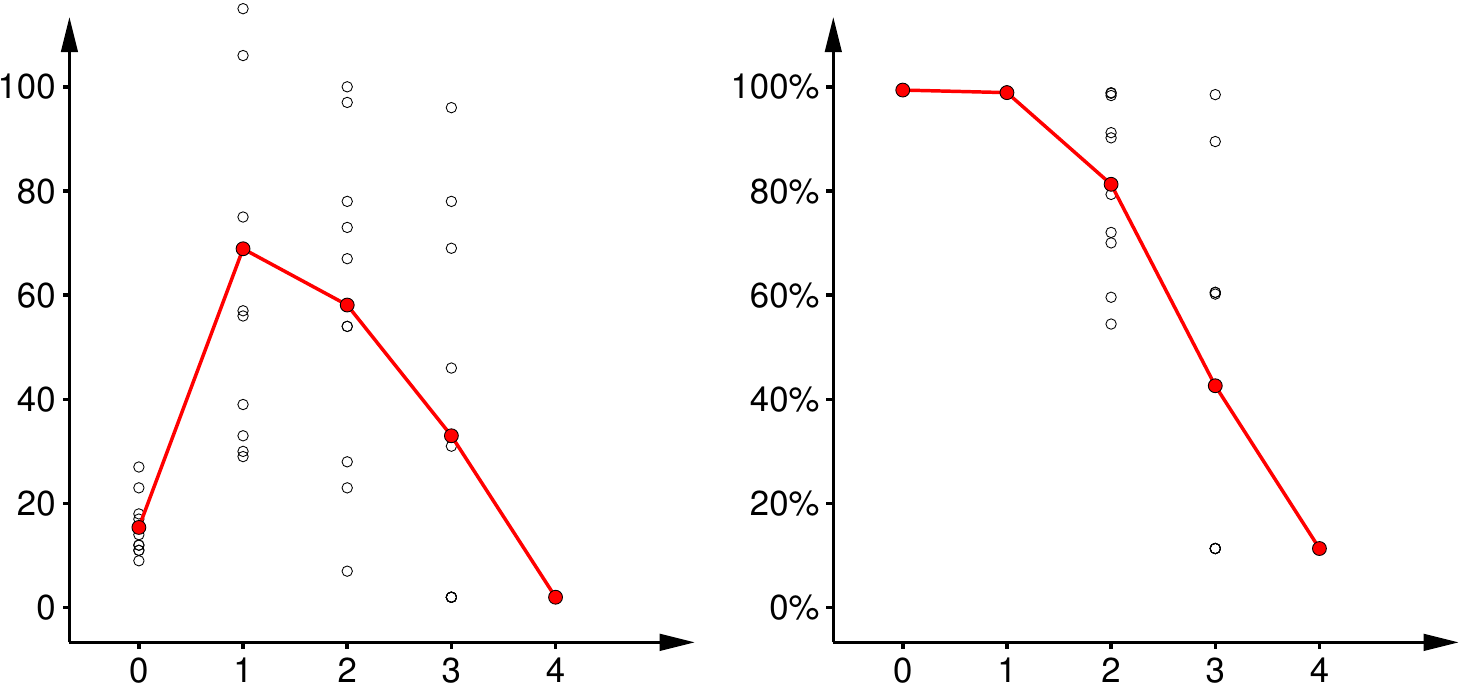}
}
\caption{\label{figure:experiment-2}
	Number of epochs (left) and resulting classification accuracy
	(right) for the handwritten digit recognition task, with
	$N \in \{0, 1, 2, 3, 4\}$ modules stacked on top of the basic MNIST
	module. The solid lines are averaged over 10 runs, individual runs
	are represented as black circles.
}
\end{center}
\end{figure}

Validation accuracies and numbers of epochs for this experiment are
provided in figure~\ref{figure:experiment-2}. It is understood that the
results are no better than in the first experiment; actually, they are
worse. With a single module on top of the basic MNIST module, the system
achieves a accuracy of nearly $99\%$ correct, however, with
significantly increased training effort compared to the basic MNIST
module and slightly worse accuracy (and with frequent failures if the
early stopping criterion is set to 10 instead of 20 epochs without
improvement of the validation error). For $N = 2$ the performance drops
significantly, in only 3 of 10 runs does the network manage to classify
all digit classes correctly. For $N = 3$, this rate drops to 1 out of 10
runs, and with $N = 4$ modules, learning fails entirely. The weights
remain very close to their initial values.

\begin{figure}
\begin{center}
{
	\begin{tabular}{cc}
	initial filters &
	$N = 0$ modules \\[0.5em]
	\includegraphics[width=0.48\textwidth]{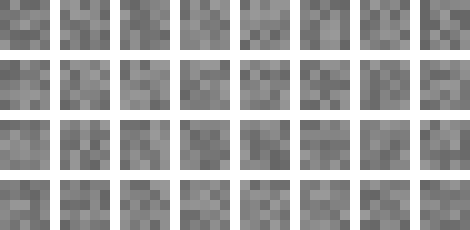} &
	\includegraphics[width=0.48\textwidth]{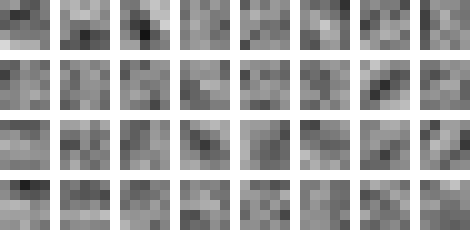} \\[0.5em]
	 &
	classification rate: 99.34\% correct \\[1.5em]
	$N = 1$ modules &
	$N = 2$ modules \\[0.5em]
	\includegraphics[width=0.48\textwidth]{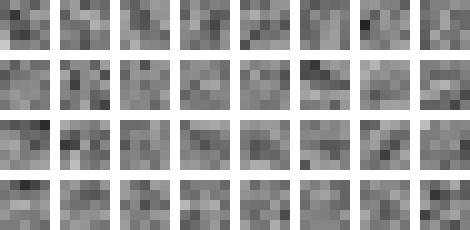} &
	\includegraphics[width=0.48\textwidth]{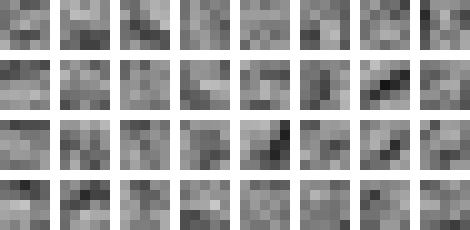} \\[0.5em]
	classification rate: 98.90\% correct &
	classification rate: 59.59\% correct \\[1.5em]
	$N = 3$ modules &
	$N = 4$ modules \\[0.5em]
	\includegraphics[width=0.48\textwidth]{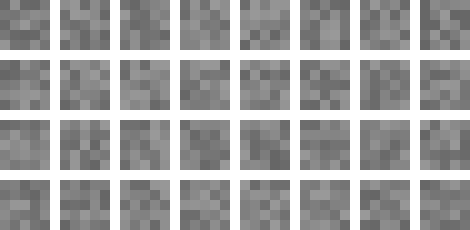} &
	\includegraphics[width=0.48\textwidth]{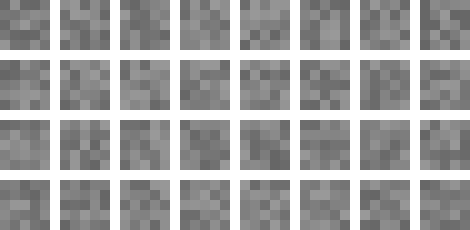} \\[0.5em]
	classification rate: 11.35\% correct &
	classification rate: 11.35\% correct
	\end{tabular}
}
\caption{\label{figure:filters}
	Visualization of the convolution filters in the first layer before
	and after training, and the corresponding validation accuracy, for
	the first out of 10 runs.
}
\end{center}
\end{figure}

Figure~\ref{figure:filters} illustrates the learned filters of the first
network layer at the end of the epoch with best validation accuracy,
i.e., the network selected by early stopping.
Various orientations are clearly recognizable for the first three
networks. All networks are initialized exactly the same, hence they
evolve similar filters. The networks with $N \geq 3$ modules fail to
evolve meaningful filters; indeed, the filter weights remain close to
their initial values.

As a control experiment, we tested the same networks with the basic
MNIST module initialized with well-tuned filters, and all further
modules initialized at random as before. The results were overall
comparable. With $N = 4$ modules on top the networks did not learn,
however, training also did not destroy the existing filters.

We find that stacking non-trivial modules on top of the basic MNIST
module results in significant slow-downs, and quickly to a complete
breakdown of end-to-end training. In this case, none of the layers is
able to learn, not even the first convolution layer, which is closest to
the data. This is easily explained by the fact that the gradient at each
node is affected by all weights during the forward pass, and by all
subsequent weights during the backwards pass. Hence, as long as none of
the components is reasonably well-trained, no component can expect to
receive a meaningful update direction, unless there is a rather broad
and hopefully straight path from random initialization to the goal
state. This is in contrast to greedy, layer-wise training, which is
entirely unaffected by later layers, and by their detrimental effect on
the gradient.

\subsection{Planning}

In the board game of RoboRally,%
\footnote{Designer: Richard Garfield; Publisher (as of 2017): Hasbro, Inc.}
in each turn a simplistic mobile robot is pre-programmed for five
movements steps in a row, in a race to reach a goal location. Most of
the fun of the game comes from the impossibility of reliable planning
due to unforeseen interactions with other robots, programmed by other
players, resulting in robots being pre-programmed with the best
intentions, but moving straight into fatal disaster. For our planning
task, we ignore these interactions and consider a simplified setup that
is more alike to a classic grid world navigation task, however, we stick
to the game element of pre-programming five moves in a row.

\begin{figure}
\begin{center}
{
	\includegraphics{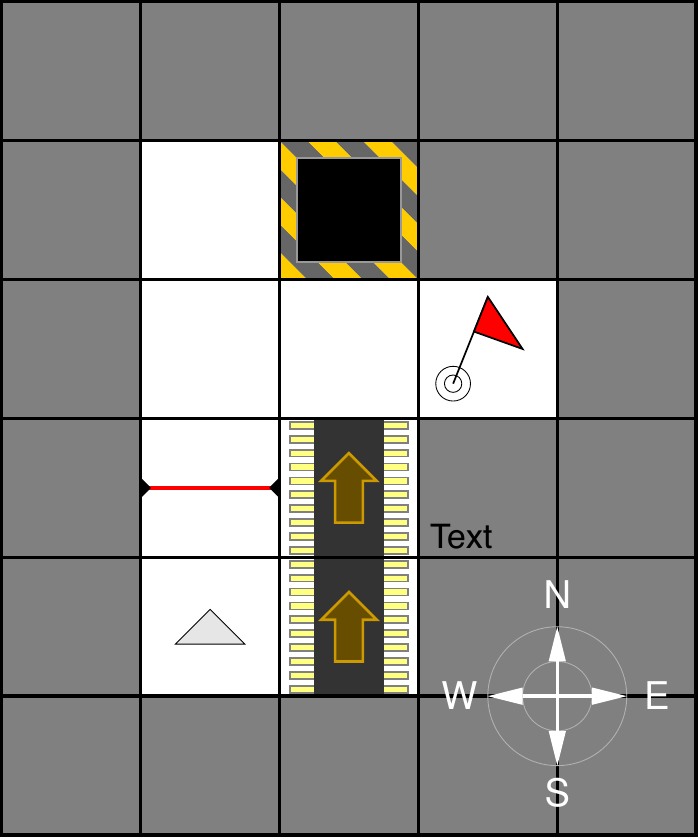}
}
\caption{\label{figure:map}
	Visualization of the RoboRally type grid-world navigation task. The
	starting state is in the lower left, with initial orientation north
	(upwards). Conveyor belts are are marked with bold arrows, the laser
	with a red line, the bottomless pit is displayed as a black square
	with a striped boundary in yellow and gray, and the goal is marked
	with a flag. The dark gray cells are walls; they cannot be entered.
	Overall the environment has 24 reachable states, including two
	states for the pit (facing north or east).
}
\end{center}
\end{figure}

The robot's state is given by its position (2D integer coordinates) and
orientation (north, south, east, west) in a grid world. It plays only a
single turn, consisting of five elementary actions chosen from the set
\textit{move\_forward}, \textit{turn\_left}, \textit{turn\_right}, and
\textit{wait}.%
\footnote{The \textit{wait} action can be useful when interacting with
  certain game elements like conveyor belts. The original game dynamics
  are more complex, adding multi-step and backwards movements, and in
  particular by executing the steps of multiple robots in a specific
  order.}
A grid cell is either a normal (empty) cell, the goal cell, a wall
(obstacle), a bottomless pit, a cell with a laser, or a conveyor belt.
Moving into a wall is a legal and available action, but does not change
the robot's state. Entering a pit cell kills the robot instantly,
entering the goal cell results in immediate success. In these cases the
robot stops moving (ignoring further actions), so the states are
terminal. However, each episode is always run for five steps and the
robot keeps collecting rewards. Hence reaching the goal quickly is
beneficial. For simplicity, in our setup all conveyor belts transport
the robot one field to the north (after its own movement), and lasers
(usually damaging the robot) have no effect other than giving negative
reward. Rewards are structured as follows: goal +10, pit -10, laser -1,
and in order to encourage exploration, the \textit{wait} action is
penalized with an ``intrinsic'' reward of -1. This is a quite simple and
actually deterministic Markov decision process (MDP). We used the map
shown in figure~\ref{figure:map}, which allows the robot to reach the
goal in only four steps (optimally: \textit{turn\_right} followed
three times by \textit{move\_forward}).

\begin{figure}
\begin{center}
{
	\includegraphics[scale=0.95]{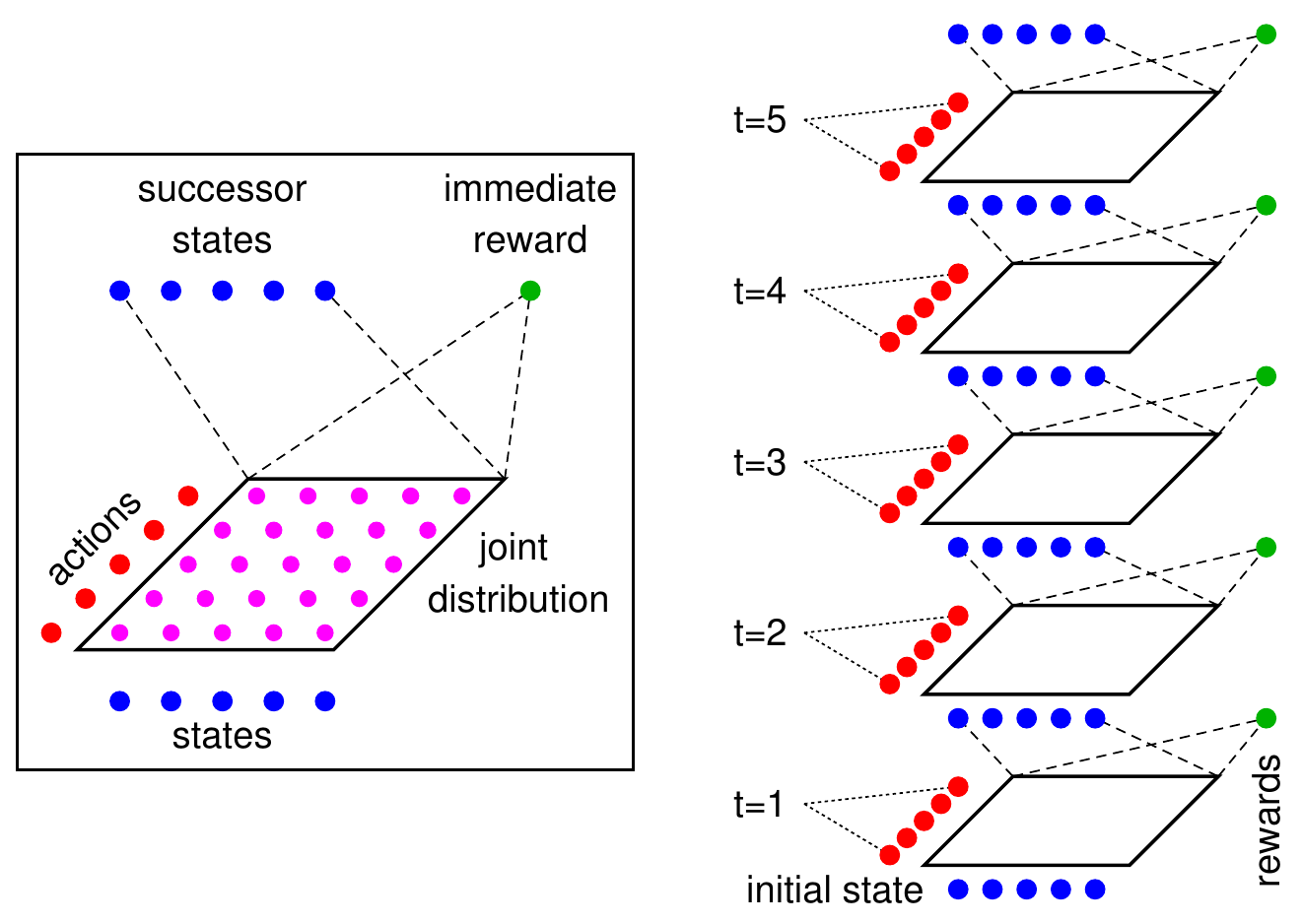}
}
\caption{\label{figure:planner}
	System architecture of the agent. Computations proceed from bottom
	to top. The forward model is shown on the left. The computation of
	the joint distribution (outer product) is hard-coded, only the
	linear read-out layers (dashed) for the successor state distribution
	and the expected immediate reward are trainable. The action selector
	consists of a softmax layer (dotted) with five inputs for the five
	time steps. The planning module is shown on the right. It applies
	action selector and forward model five times. The action selector is
	best understood as encoding five independent action distributions.
	The forward model is used five times in a row, hence the planner can
	be thought of as containing five copies of the network with shared
	weights. The initial state and the time steps are provided as
	(constant) inputs, so that all adjustable quantities are represented
	as network weights.
}
\end{center}
\end{figure}

Playing this game successfully and even optimally is not hard at all,
since there are only $4^5 = 1024$ possible action sequences, $25$ of
which reach the goal, and the environment is deterministic. Our agent
proceeds by planning its actions based on a forward model, which by
itself is a neural network and must be learned from interactions. Given
a state-action pair $(s, a)$ with $s$ and $a$ each one-hot-encoded
(representing explicit probabilities, the outer product representing the
joint distribution acts as the actual input), the forward model maps to
a probability distribution over successor states. It is implemented by a
fully connected network layer with linear activation and probability
simplex constraints. Expected immediate rewards are modeled analogously
with a linear layer without any constraints. This model is extremely
simple in nature; it is essentially linear. The model can capture the
true transition and reward structure of the MDP exactly, and it is
suitable for propagating distributions over states and actions
arbitrarily far into the future. This property makes the model suitable
for planning, e.g., by executing mental trials.
Of course, the predictions will be off as long as the model is not yet
well trained. The actions are encoded as probability distributions,
realized as a softmax layer receiving five one-hot encoded ``time step''
inputs. The overall agent consists of two trainable modules, the world
model mapping state $s$ and action $a$ to successor state $s'$ and
reward $r$, and the programming module (action selector module), mapping
time $t \in \{1, 2, 3, 4, 5\}$ to an action $a$ from the action set
defined above. The full system is displayed in
figure~\ref{figure:planner}.

The system is trained in an e2e manner for 10,000 episodes. In each
step, an episode is planned. The plan is refined by a gradient step on
the programming module, based on the total reward predicted by the
forward model. I.e., the agent improves its policy based on its current
world model. Then an actual action sequence is sampled from the updated
action distribution (the actual action in the RoboRally game), and the
episode is executed in the environment. Note that this setup differs
from standard reinforcement learning, since the agent essentially
performs only a single action with a ``reactive'' policy, consisting of
five sub-actions, which must be pre-programmed and hence planned
upfront. Then the world model module is updated based on the observed
successor states and rewards.%
\footnote{Our learning system is related to value iteration networks
  \cite{tamar2016value}, however, in our case the term \emph{planning}
  actually refers to performing a full ``mental trial'' based on a
  learned forward model.}

This setup is obviously not the most efficient possible for the task at
hand, which can of course be solved easily, e.g., by tabular
reinforcement learning, or simply by brute force enumeration of all
action sequences, or assuming possible stochasticity, by multi-armed
bandit algorithms. Therefore it must be emphasized that this is a toy
experiment, the only purpose of which is to demonstrate the effects of
unmodeled dependencies between learning modules. Our learning system
indeed has the property of interest, namely a non-trivial mutual
dependency of the two learning modules: the action selector depends on
the quality of the world model for correct forward planning, and the
world model depends on the action selector to perform sufficient
exploration so that it gets to see the relevant parts of the state
space.

Despite the simplicity of the task, learning both modules in the above
described intertwined manner works only in 47 out of 100 runs. In the
other 53 runs the system quickly converges to a local optimum,
consisting of turning movements, since they avoid negative rewards. This
task is on the edge of the manageable difficulty; even slightly more
complex tasks result in near certain failure. This is despite the facts
that both modules can represent the optimal solution exactly (the
solution is realizable), the time horizon is rather small, the
environment consists of only 24 states with four actions, and learning
any of the modules in isolation works flawlessly: learning the MDP from
random actions works fine, and so does learning the optimal policy from
the exact MDP or a pre-trained forward model.%
\footnote{Care must be taken not to insert too many pits, since
  otherwise the chance of dying during exploration exceeds that or
  reaching the goal. That turns ``not moving'' into a local optimum,
  which is easily reachable from the initial random policy. An explicit
  exploration strategy would be needed to overcome this problem.}
This result suggests a simple solution: first train the world model till
convergence based on random actions, then train the action selector.
This decomposition of the training process is exactly in line with the
problem decomposition underlying the network design. Of course, in its
simple form this solution is dissatisfactory since random exploration is
often inefficient, but it is a simplistic demonstration showing that a
learning process organized along the network structure can indeed be a
viable solution.

\section{Conclusion}

We have demonstrated that end-to-end learning can be very inefficient
for training neural network models composed of multiple non-trivial
modules. End-to-end learning can even break down entirely; in the
worst case none of the modules manages to learn. In contrast, each
module is able to learn if the other modules are already trained
and their weights frozen. This suggests that training of complex
learning machines should proceed in a \emph{structured} manner, training
simple modules first and independent of the rest of the network.

Our example problems are necessarily somewhat contrived.
Considering neural networks designed for solving real tasks, whether the
limits of end-to-end will show up in the foreseeable future or not
remains to be seen. At this point we simply want to raise awareness for
the existence of limitations. To overcome these problems in a principled
manner, we believe that new structured learning paradigms are needed,
which should be in line with the network structure, and which may or may
not contain greedy learning and end-to-end learning as techniques. We
are convinced that such structured learning paradigms would allow us to
push the boundaries of training complex learning systems beyond the
current state-of-the-art.

\section*{Acknowledgments}
I would like to thank Oswin Krause for helpful discussions and comments.

\bibliographystyle{plain}

\begin{thebibliography}{10}

\bibitem{abbeel2004apprenticeship}
P.~Abbeel and A.~Ng.
\newblock Apprenticeship learning via inverse reinforcement learning.
\newblock In {\em International Conference on Machine learning}, page~1. ACM,
  2004.

\bibitem{keras}
F.~Chollet.
\newblock Keras.
\newblock \url{https://github.com/fchollet/keras}, 2015.

\bibitem{cirecsan2011committee}
D.~Cire{\c{s}}an, U.~Meier, J.~Masci, and J.~Schmidhuber.
\newblock A committee of neural networks for traffic sign classification.
\newblock In {\em International Joint Conference on Neural Networks (IJCNN)},
  pages 1918--1921. IEEE, 2011.

\bibitem{collobert2011natural}
R.~Collobert, J.~Weston, L.~Bottou, M.~Karlen, K.~Kavukcuoglu, and P.~Kuksa.
\newblock Natural language processing (almost) from scratch.
\newblock {\em Journal of Machine Learning Research}, 12(Aug):2493--2537, 2011.

\bibitem{deng2009imagenet}
J.~Deng, W.~Dong, R.~Socher, Li-Jia Li, Kai L., and L.~Fei-Fei.
\newblock {Imagenet: A large-scale hierarchical image database}.
\newblock In {\em IEEE Conference on Computer Vision and Pattern Recognition},
  pages 248--255, 2009.

\bibitem{tensorflow}
M.~Abadi et~al.
\newblock {TensorFlow: Large-Scale Machine Learning on Heterogeneous
  Distributed Systems}.
\newblock Technical Report arXiv:1603.04467, arxiv.org, 2016.

\bibitem{graves2014neural}
A.~Graves, G.~Wayne, and I.~Danihelka.
\newblock Neural turing machines.
\newblock Technical Report arXiv:1410.5401, arxiv.org, 2014.

\bibitem{graves2016hybrid}
A.~Graves, G.~Wayne, M.~Reynolds, T.~Harley, I.~Danihelka,
  A.~Grabska-Barwi{\'n}ska, S.~Colmenarejo~G{\'o}mez, E.~Grefenstette,
  T.~Ramalho, and J.~Agapiou.
\newblock Hybrid computing using a neural network with dynamic external memory.
\newblock {\em Nature}, 538(7626):471--476, 2016.

\bibitem{he2016deep}
K.~He, X.~Zhang, S.~Ren, and J.~Sun.
\newblock Deep residual learning for image recognition.
\newblock In {\em IEEE Conference on Computer Vision and Pattern Recognition},
  pages 770--778, 2016.

\bibitem{hinton2006fast}
G.E. Hinton, S.~Osindero, and Y.-W. Teh.
\newblock A fast learning algorithm for deep belief nets.
\newblock {\em Neural computation}, 18(7):1527--1554, 2006.

\bibitem{caffe}
Y.~Jia, E.~Shelhamer, J.~Donahue, S.~Karayev, J.~Long, R.~Girshick,
  S.~Guadarrama, and T.~Darrell.
\newblock {Caffe: Convolutional architecture for fast feature embedding}.
\newblock In {\em International Conference on Multimedia}, pages 675--678. ACM,
  2014.

\bibitem{johnson2013accelerating}
R.~Johnson and T.~Zhang.
\newblock {Accelerating Stochastic Gradient Descent using Predictive Variance
  Reduction}.
\newblock In {\em Advances in Neural Information Processing Systems (NIPS)},
  pages 315--323, 2013.

\bibitem{kingma2014adam}
D.~Kingma and J.~Ba.
\newblock {Adam: A method for stochastic optimization}.
\newblock Technical Report arXiv:1412.6980, arxiv.org, 2014.

\bibitem{alexnet}
A.~Krizhevsky, I.~Sutskever, and G.E. Hinton.
\newblock {ImageNet Classification with Deep Convolutional Neural Networks}.
\newblock In F.~Pereira, C.~J.~C. Burges, L.~Bottou, and K.~Q. Weinberger,
  editors, {\em Advances in Neural Information Processing Systems}, pages
  1097--1105. 2012.

\bibitem{lecun2005offroad}
Y.~LeCun, U.~M{\"{u}}ller, J.~Ben, E.~Cosatto, and B.~Flepp.
\newblock Off-road obstacle avoidance through end-to-end learning.
\newblock In {\em Neural Information Processing Systems}, pages 739--746, 2005.

\bibitem{mirowski2016learning}
P.~Mirowski, R.~Pascanu, F.~Viola, H.~Soyer, A.~Ballard, A.~Banino, M.~Denil,
  R.~Goroshin, L.~Sifre, and K.~Kavukcuoglu.
\newblock Learning to navigate in complex environments.
\newblock Technical Report arXiv:1611.03673, arxiv.org, 2016.

\bibitem{mnih2015human}
V.~Mnih, K.~Kavukcuoglu, D.~Silver, A.A. Rusu, J.~Veness, M.G. Bellemare,
  A.~Graves, M.~Riedmiller, A.K. Fidjeland, and G.~Ostrovski.
\newblock Human-level control through deep reinforcement learning.
\newblock {\em Nature}, 518(7540):529--533, 2015.

\bibitem{polyak1963gradient}
B.T. Polyak.
\newblock Gradient methods for the minimisation of functionals.
\newblock {\em USSR Computational Mathematics and Mathematical Physics},
  3(4):864--878, 1963.

\bibitem{ring1994continual}
M.~Ring.
\newblock {\em Continual Learning in Reinforcement Environments}.
\newblock PhD thesis, University of Texas at Austin, 1994.

\bibitem{schmidhuber2004optimal}
J.~Schmidhuber.
\newblock Optimal ordered problem solver.
\newblock {\em Machine Learning}, 54(3):211--254, 2004.

\bibitem{schmidhuber2010formal}
J.~Schmidhuber.
\newblock Formal theory of creativity, fun, and intrinsic motivation.
\newblock {\em IEEE Transactions on Autonomous Mental Development},
  2(3):230--247, 2010.

\bibitem{schmidhuber2015deep}
J.~Schmidhuber.
\newblock {Deep learning in neural networks: An overview}.
\newblock {\em Neural networks}, 61:85--117, 2015.

\bibitem{schmid2013minimizing}
M.~Schmidt, N.~Le~Roux, and F.~Bach.
\newblock {Minimizing Finite Sums with the Stochastic Average Gradient}.
\newblock Technical Report arXiv:1309.2388, arxiv.org, 2013.

\bibitem{silver2016mastering}
D.~Silver, A.~Huang, C.J. Maddison, A.~Guez, L.~Sifre, G.~Van Den~Driessche,
  J.~Schrittwieser, I.~Antonoglou, V.~Panneershelvam, and M.~Lanctot.
\newblock Mastering the game of {Go} with deep neural networks and tree search.
\newblock {\em Nature}, 529(7587):484--489, 2016.

\bibitem{szegedy2015going}
C.~Szegedy, W.~Liu, Y.~Jia, P.~Sermanet, S.~Reed, D.~Anguelov, D.~Erhan,
  V.~Vanhoucke, and A.~Rabinovich.
\newblock Going deeper with convolutions.
\newblock In {\em IEEE Conference on Computer Vision and Pattern Recognition},
  pages 1--9, 2015.

\bibitem{tamar2016value}
A.~Tamar, S.~Levine, P.~Abbeel, Y.~Wu, and G.~Thomas.
\newblock Value iteration networks.
\newblock In {\em Advances in Neural Information Processing Systems}, pages
  2146--2154, 2016.

\bibitem{theano}
{Theano Development Team}.
\newblock Theano: A {Python} framework for fast computation of mathematical
  expressions.
\newblock Technical Report arXiv:1605.02688, arxiv.org, 2016.

\bibitem{matconvnet}
A.~Vedaldi and K.~Lenc.
\newblock {MatConvNet -- Convolutional Neural Networks for MATLAB}.
\newblock In {\em Proceeding of the {ACM} Int. Conf. on Multimedia}, 2015.

\bibitem{vincent2008extracting}
P.~Vincent, H.~Larochelle, Y.~Bengio, and P.-A. Manzagol.
\newblock Extracting and composing robust features with denoising autoencoders.
\newblock In {\em International Conference on Machine Learning}, pages
  1096--1103. ACM, 2008.

\bibitem{zeiler2012adadelta}
M.D. Zeiler.
\newblock {ADADELTA}: an adaptive learning rate method.
\newblock Technical Report arXiv:1212.5701, arxiv.org, 2012.

\end{thebibliography}

\end{document}